\documentclass[letterpaper, 10 pt, conference]{ieeeconf}  
\usepackage{graphicx}
\usepackage{amsmath}      
\usepackage{amssymb}
\usepackage{booktabs}
\usepackage{array}
\usepackage{arydshln}
\usepackage{hyperref}
\usepackage{comment}
\usepackage{placeins}
\usepackage{bm}       
\usepackage{booktabs, multirow, siunitx}    
\usepackage{xcolor}  
\usepackage[T1]{fontenc}
\hypersetup{
	colorlinks=true,   
	linkcolor=blue,    
	citecolor=blue,    
	urlcolor=blue,     
	filecolor=blue,    
	anchorcolor=blue   
}

\IEEEoverridecommandlockouts                              

\title{\LARGE \bf
D-GVIO: A Buffer-Driven and Efficient Decentralized GNSS-Visual-Inertial State Estimator for Multi-Agent Systems
}

\author{Yarong Luo$^{1}$, Wentao Lu$^{2}$, Chi Guo$^{1}$ and Ming Li$^{1}$
	\thanks{$^{1}$Yarong Luo, Chi Guo and Ming Li are with the School of Robotics, Wuhan University, Wuhan 430072, China. \{yarongluo,~guochi,~liwhuer\}@whu.edu.cn}%
	\thanks{$^{2}$Wentao Lu is with the School of Electronic Information of Wuhan
		University, Hubei Luojia Laboratory, Wuhan 430072, China. wentaolu@whu.edu.cn}%
		\thanks{This work is supported in part by the National Natural Science Foundation  of China under Grant 42404025 (Corresponding author: Chi Guo).}
	\thanks{Our code is publicly available: \url{https://github.com/braveryyyy/D-GVIO}.}%
}

\begin{document}

\maketitle
\thispagestyle{empty}
\pagestyle{empty}

\begin{abstract}
Cooperative localization is essential  for swarm applications like collaborative exploration and search-and-rescue missions. However, maintaining real-time capability, robustness, and computational efficiency on resource-constrained platforms presents significant challenges. To address these challenges, we propose D-GVIO, a buffer-driven and fully decentralized GNSS-Visual-Inertial Odometry (GVIO) framework that leverages a novel buffering strategy to support efficient and robust distributed state estimation. The proposed framework is characterized by four core mechanisms. Firstly, through covariance segmentation, covariance intersection and buffering strategy, we modularize propagation and update steps in distributed state estimation, significantly reducing computational and communication burdens. Secondly, the left-invariant extended Kalman filter (L-IEKF) is adopted for information fusion, which exhibits superior state estimation performance over the traditional extended Kalman filter (EKF) since its state transition matrix is independent of the system state. Thirdly, a buffer-based re-propagation strategy is employed to handle delayed measurements efficiently and accurately by leveraging the L-IEKF, eliminating the need for costly re-computation. Finally, an adaptive buffer-driven outlier detection method is proposed to dynamically cull GNSS outliers, enhancing robustness in GNSS-challenged environments.
\end{abstract}

\section{INTRODUCTION}
In recent years, multi-agent systems have demonstrated great potential in a variety of applications, including collaborative autonomous exploration \cite{gao2022meeting}, \cite{zhou2023racer} and search-and-rescue missions \cite{berger2015innovative},  \cite{li2023collaborative}. Robust collaborative localization enables agent teams to operate autonomously in unstructured and dynamic scenarios. A fundamental requirement for real-world multi-agent deployments is distributed state estimation, in which each agent estimates its own state from local observations.

\begin{figure}[tbp]
	\centering
	\includegraphics[width=0.45\textwidth]{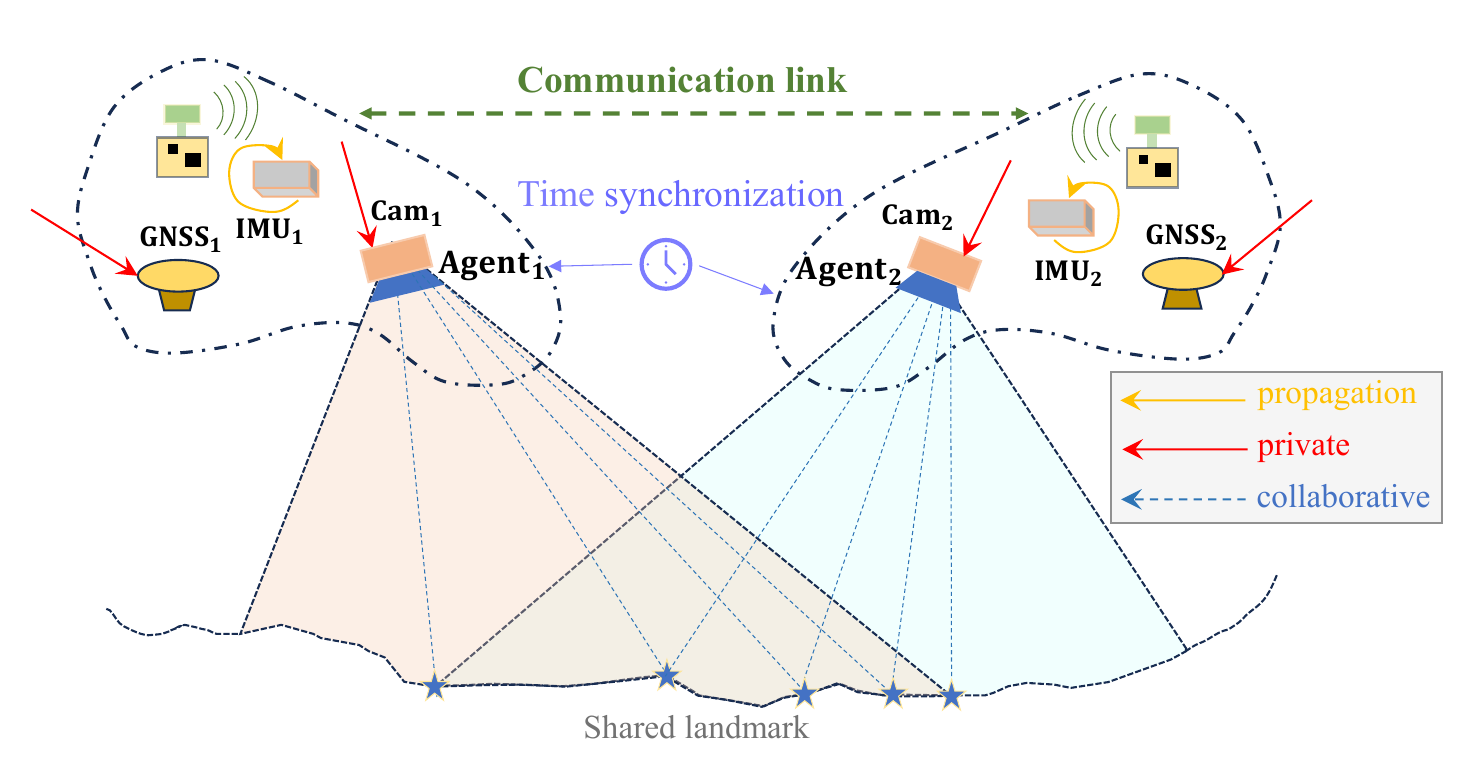}
	\caption{Decentralized GVIO system with two agents ($\textrm{Agent}_1$, $\textrm{Agent}_2$). Each agent runs a local filter for propagation and private updates using IMU, GNSS, and camera data. Collaborative updates occur when both agents observe common landmarks via feature matching.}
	\label{frame}
\end{figure}

Currently, in outdoor scenes, Global Navigation Satellite System (GNSS), particularly Real-Time Kinematic (RTK), is widely used for estimating the agent's position, as GNSS provides global position without accumulating drift over time \cite{song2024gps}. In contrast, in GNSS-denied environments such as indoor scenes, visual-inertial odometry (VIO) \cite{10611284} or LiDAR-inertial odometry (LIO) \cite{11127793} are much preferred. Previous studies \cite{song2024gps}, \cite{10477234} have shown that combining GNSS positioning with VIO enables high-precision localization while maintaining low-drift performance at the global scale. Therefore, the integration of GNSS and VIO offers an effective solution for achieving accurate and robust distributed state estimation in multi-agent systems (see Fig. \ref{frame}).

However, applying distributed state estimation to multi-agent systems remains challenging. As the number $n$ of sensors per agent increases, the computational burden of naive estimators grows cubically~\cite{9286578}, imposing heavy computational loads and degrading real-time performance, especially on resource-constrained platforms. Additionally, maintaining cross-covariance with other agents leads to communication and memory burdens that grow linearly with the number of encountered agents \cite{9561203}. Thus, methods like distributed approximated cross-covariance \cite{luft2018recursive} are not well-suited for large-scale swarms and long-duration missions involving frequent encounters. Moreover, information fusion from different sensors remains challenging due to varying sampling rates and computational delays. Communication and computational delays between agents are also inevitable \cite{9445652}. These issues can lead to delayed measurements, which in turn may trigger extensive re-computations to maintain filter consistency.

To this end, we propose a decentralized GVIO system. This system features low computational and memory burdens, effective handling of delayed measurements, and robustness to unreliable GNSS observations. These characteristics make D-GVIO well-suited for resource-constrained platforms. More precisely, our contributions are:
\begin{itemize}  
	\item A buffer-driven decentralized collaborative GVIO system enabling efficient and robust distributed state estimation. This system features low computational and memory costs, and integrates an efficient delayed measurement handling strategy using invariant extended Kalman filter (IEKF).
	
	\item An efficient GNSS outlier detection and culling strategy. Unlike traditional chi-square test approaches that evaluate Mahalanobis distance, this system employs a buffer-based method that detects outliers by comparing GNSS velocities against an entropy-adaptive threshold derived from VIO kinematics and the Boltzmann criterion.
	
	\item Comprehensive experiments on open-source and real-world datasets. Experimental results demonstrate the efficiency and reliability of D-GVIO. Furthermore, in order to support the community's development, we have open-sourced our code and datasets on GitHub.
\end{itemize}

\section{RELATED WORK}
{
	In recent years, cooperative localization for multi-agent systems has received significant attention. Existing approaches in cooperative localization can be divided into centralized and decentralized \cite{9844233}. The former relies on a central unit acting as a server, where all agents upload their sensor data for fusion. In contrast, the latter performs data fusion locally on each agent, which requires efficient fusion mechanisms and minimal data exchange \cite{10816004}, \cite{DUBOIS2022103933}. Although a centralized system is easier to implement, it relies on a central entity that must always be reachable and fault-free. Therefore, this work focuses on the decentralized approach, which is more robust and scalable.
	
	In decentralized framework, using sensors such as cameras, inertial measurement unit (IMU), or LiDAR to perform collaborative simultaneous localization and mapping (SLAM) for multi-agent systems is an intuitive approach. 
    For cooperative SLAM in large-scale swarms, Schmuck et al. proposed COVINS \cite{schmuck2021covins}, which leverages the dynamic redundancy pruning strategy, and efficient pose graph management to achieve sub-decimeter localization accuracy in multi-agent systems.
    However, this system relies on pose graph optimization, incurring high computational costs and poor real-time performance. Consequently, filter-based approaches become preferable for resource-limited platforms. 
	Zhu et al. proposed Swarm-LIO \cite{10161355}, a decentralized filter-based swarm LIO system achieving centimeter-level accuracy, and its successor Swarm-LIO2 \cite{10816004}, which improves scalability by using dynamic state compression and time compensation to limit state dimension growth with swarm size.
	The aforementioned studies for cooperative localization rely on sensors such as IMU, LiDAR, and camera. Nevertheless, these sensors cannot provide global positioning, leading to drift accumulation \cite{10161355}. 
	Thus, GNSS integration, particularly with high-precision RTK positioning, becomes crucial in GNSS-available environments. Meanwhile, in challenged scenarios such as tunnels and dense urban areas, efficient outlier detection and robust measurement weighting strategies are required to handle degraded and unreliable observations.
	
	Distributed state estimation on resource-limited platforms is vital for cooperative localization. Polizzi et al. \cite{9844233} proposed a decentralized VIO using a vector of locally aggregated descriptors (VLAD) \cite{7025567} based request-response strategy for low bandwidth. However, the high-dimensional covariance matrix required during propagation step results in a great computational burden. Addressing this, Bromme et al. \cite{9286578} proposed a filtering-based multi-sensor fusion method using covariance segmentation. Jung et al. \cite{9197178} then extended this to multi-agent systems, enabling efficient distributed state estimation. This system minimizes communication and computation by only exchanging data during joint observations. However, it scales poorly to large swarms as maintaining inter-agent dependencies incurs memory and communication burdens that grow linearly with the number of encountered agents.
	
	Moreover, handling delayed measurements from communication and computation delays  remains challenging. Existing re-computation approaches \cite{9286578}, \cite{9844233} incur high computational burdens when high-frequency sensors are used. For multi-agent systems, Tasooji et al. \cite{9888790} proposed a decentralized event-triggered method to mitigate accuracy loss induced by delayed measurements, but this method requires state augmentation, which increases the computational burdens. To efficiently handle delayed measurements, Alexander \cite{alexander1991state} proposed a re-propagation method that propagates the covariance and mean of error-state from the measurement occurrence moment to the update completion moment. However, the traditional EKF-based re-propagation method cannot avoid accuracy degradation, since its state transition matrix depends on the estimated state and thus becomes inaccurate when the estimated state deviates from the true state. Thus, it is meaningful to investigate how to process delayed measurements rapidly while maintaining state estimation accuracy.
}
\section{METHODOLOGY}
\subsection{Overview}
\begin{figure*}[t!]  
		\centering
		\includegraphics[width=0.95\textwidth]{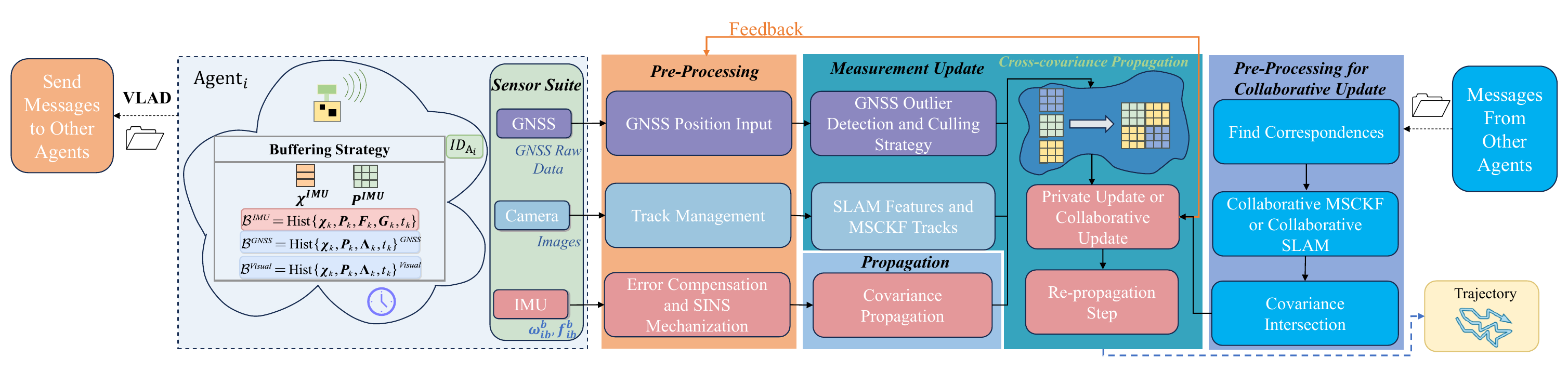}
		\caption{Block diagram of the proposed system. Each agent maintains local SLAM features and MSCKF tracks for measurement updates. When receiving messages from other agents, the system finds feature correspondences and performs collaborative MSCKF or SLAM updates.}
		\label{system_frame}
	\end{figure*}

	This work considers a system of $n$ agents, each equipped with an IMU, camera, and GNSS, maintaining a local filter and capable of inter-agent communication (e.g., via WiFi). A tightly-coupled approach is employed for camera-IMU fusion due to its superior accuracy and robustness \cite{8258858}. Conversely, a loosely-coupled approach is used for GNSS-IMU fusion to facilitate decoupling during GNSS outages.
	As illustrated in Fig.~\ref{system_frame}, the proposed D-GVIO is organized into three major modules:  
	(i)~The pre-processing module handles GNSS position input, track management, and IMU error compensation with SINS mechanization;  
	(ii)~The measurement update module performs GNSS and visual information fusion efficiently and robustly through private updates, supported by outlier detection, cross-covariance propagation, and re-propagation for delayed measurements;  
	(iii)~The collaborative update module establishes feature correspondences across agents and performs either collaborative multi-state constraint Kalman filter (MSCKF) or SLAM updates, with consistency ensured via covariance intersection (CI) algorithm \cite{julier1997non}. Due to the space constraints, the specific procedures for collaborative MSCKF and SLAM updates can be found in \cite{9844233}.
	\subsection{Reference Frames and Notations}
		\begin{figure}[htbp]
		\centering
		\includegraphics[width=0.4\textwidth]{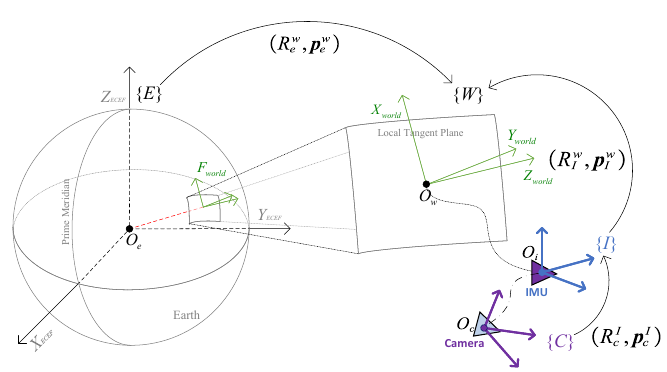}
		\caption{Coordinate frames in the GVIO system and their transformations
			between each other. The transformation of frame $\{C\}$ into the frame $\{I\}$ is $\left(\bm{R}^I_c,\bm{p}^I_c\right)$ and the transformation of frame $\{I\}$ into the frame $\{W\}$ is $\left(\bm{R}^w_I,\bm{p}^w_I\right)$.}
		\label{coordinate}
	\end{figure}
	The coordinate frames used in this work are shown in the Fig. \ref{coordinate}. $\{W\}$ denotes the world frame. $\{E\}$ represents the Earth-Centered Earth-Fixed (ECEF) frame, which is the reference frame of GNSS position measurements. These measurements can be transformed into the world frame $\{W\}$ using the rotation matrix $\bm{R}^w_e$ and translation vector $\bm{p}^w_e$. $\{I\}$ represents the IMU frame rigidly attached to the IMU, and $\{C\}$ denotes the camera frame rigidly attached to the camera.
	
	In this work, the notation $\left(\bullet\right)^w$ represents a quantity expressed in the frame $\{W\}$. The velocity and position of the IMU w.r.t. the frame $\{W\}$ are expressed as $\bm{v}^w_I$ and $\bm{p}^w_I$, respectively. The notation $\bm{R}^w_I$ represents the rotation matrix which rotates a vector $\bm{x}^I$ defined in the frame $\{I\}$ to a vector $\bm{x}^w=\bm{R}^w_I\bm{x}^I$ defined in the $\{W\}$ frame. Furthermore, $[\bullet]\times$ denotes the skew-symmetric matrix of a three-dimensional vector, while $[\bullet]^{\mathrm{T}}$ is used to represent the transpose of a matrix.
\subsection{State Definition and Information Fusion via IEKF}
{\label{suba}
	In D-GVIO, the system state $\bm{\chi}$ of each agent can be divided into IMU-related core states $\bm{\chi}_I$, visual-related states $\bm{\chi}_v$, and GNSS-related states $\bm{\chi}_g$
	\begin{equation}
		\bm{\chi}=\left(\bm{\chi}_I,\bm{\chi}_v,\bm{\chi}_g\right)=\left(
		{\bm{R}_I^w},{\bm{v}_I^w}, {\bm{p}_I^w}, {\bm{b}_g},{\bm{b}_a},
		\bm{\chi}_v,
		\bm{\chi}_g
		\right)
		\label{eq1}
	\end{equation}
	\begin{equation}
		\bm{\chi}_v=\left(
		\left({\bm{R}_{c_1}^w},{\bm{p}_{c_1}^w}\right)... \left({\bm{R}_{c_M}^w},{\bm{p}_{c_M}^w}\right), {\bm{f}_1}... {\bm{f}_N}
		\right)
		\label{eq3}
	\end{equation}
	where $\bm{\chi}_g=\bm{l}_b$ is the arm lever between IMU center and the GNSS antenna phase center; $\bm{\chi}$ evolves on a product manifold $\mathcal M=\mathbb{SE}_2\left(3\right)\times\left(\mathbb{R}^3\right)^2\times\left(\mathbb{SO}\left(3\right)\times\mathbb{R}^3\right)^{M}\times\left(\mathbb{R}^3\right)^N\times\mathbb{R}^3$. $\left({\bm{R}_I^w},{\bm{v}_I^w}, {\bm{p}_I^w}\right)\in\mathbb{SE}_2\left(3\right)$ is the extended pose and $\left(\bm{b}_g,\bm{b}_a\right)\in\left(\mathbb{R}^3\right)^2$ are the gyroscope and accelerometer biases, respectively. The visual-related states include the sliding window states and the feature states. The sliding window states include the poses $\left(\bm{R}^w_{c},\bm{p}^w_c\right)\in\mathbb{SO}\left(3\right)\times\mathbb{R}^3$ of the camera frame at the past $M$ image time instances. The feature states include the inverse depth parameterization of landmarks $\bm{f}_i$ with $i=1,...,N$ \cite{9844233}. (\ref{error}) defines the $(18+3N+6M)$-dimensional  logarithmic error on the manifold $\mathcal{M}$ based on IEKF.
	\begin{equation}
		\resizebox{0.90\hsize}{!}{$
		\begin{aligned}
		\bm{\eta}&=\left[\begin{array}{ccc}
			\bm{R}^I_w&-\bm{R}^I_w\bm{v}^w_I&-\bm{R}^I_w\bm{p}^w_I\\
			\mathbf{0}_{1\times3}&1&0\\\mathbf{0}_{1\times3}&0&1
		\end{array}\right]\left[\begin{array}{ccc}
		\hat{\bm{R}}^w_I&\hat{\bm{v}}^w_I&\hat{\bm{p}}^w_I\\
		\mathbf{0}_{1\times3}&1&0\\\mathbf{0}_{1\times3}&0&1
		\end{array}\right]\\
		&=\left[\begin{array}{ccc}
			\bm{R}^I_w\hat{\bm{R}}^w_I&\bm{R}^I_w\hat{\bm{v}}^w_I-\bm{R}^I_w\bm{v}^w_I&\bm{R}^I_w\hat{\bm{p}}^w_I-\bm{R}^I_w\bm{p}^w_I\\
			\mathbf{0}_{1\times3}&1&0\\\mathbf{0}_{1\times3}&0&1
		\end{array}\right]\\
		&\delta\bm{b}_g=\hat{\bm{b}}_g-\bm{b}_g,\delta\bm{b}_a=\hat{\bm{b}}_a-\bm{b}_a,\mathrm{exp}\left(\bm{\phi}^c\times\right)=\bm{R}^{c_i}_w\hat{\bm{R}}^w_{c_i}\\
		&\delta\bm{p}^w_c=\hat{\bm{p}}^w_c-\bm{p}^w_c,\delta\bm{f}_i=\hat{\bm{f}}_i-\bm{f}_i, \delta\bm{l}_b=\hat{\bm{l}}_b-\bm{l}_b
	\end{aligned}
	\label{error}
	$}
	\end{equation}
	where $\bm{\eta}$ is the left invariant error; $\delta\bm{b}_g$ and $\delta\bm{b}_a$ are the errors of gyroscope and accelerometer biases, respectively; $\mathrm{exp}\left(\cdot\right)$ denotes the exponential map of the group $\mathbb{SE}_2\left(3\right)$; $\bm{\phi}^c$ is the attitude error of the camera expressed in the form of a rotation vector; $\delta\bm{p}^w_c$ is the position error of the camera; $\delta\bm{f}_i$ and $\delta\bm{l}_b$ are the errors of the inverse depth parameterization of landmarks and arm lever, respectively.
	
	The IEKF \cite{iekf} is used for information fusion from different sensors and agents. The discrete propagation step at time $t_k$ is as follows:
	\begin{equation}
		\begin{aligned}
			\hat{\bm{\chi}}_{k|k-1}=f\left(\hat{\bm{\chi}}_{k-1|k-1},u_k\right)
		\end{aligned}
	\end{equation}
	\begin{equation}
		\begin{aligned}
			\bm{P}_{k|k-1}=&\bm{F}_{k}\bm{P}_{k-1|k-1}{\bm{F}}^{\mathrm{T}}_{k}+\bm{G}_{k}\bm{Q}_{k}\bm{G}^{\mathrm{T}}_{k}		
		\end{aligned}
	\end{equation}
	where $\hat{\bm{\chi}}_{k|k-1}$ is \textit{a priori} estimate with covariance $\bm{P}_{k|k-1}$; $f\left(\cdot\right)$ is the process function; $u_k$ is the control input (e.g. inertial measurements); $\bm{F}_{k}$ is the discrete state transition matrix; $\bm{Q}_{k}$ is the covariance of the process noise; $\bm{G}_{k}$ is the process noise matrix. It is noteworthy that the propagation step requires propagating the core states covariance and cross-covariance between the core states and the measurement sensor states.
	
	The observation information is fused into the estimated state through the update steps \cite{10232373}:
	\begin{equation}
		\resizebox{0.7\hsize}{!}{$
		\bm{K}_k=\bm{P}_{k|k-1}{\bm{H}_k}^{\mathrm{T}}\left(\bm{N}_k+\bm{H}_k^{\mathrm{T}}\bm{P}_{k|k-1}\bm{H}_k\right)^{-1}
		\label{gain}$}
	\end{equation}
	\begin{equation}
			\resizebox{0.7\hsize}{!}{$
		\begin{aligned}
			\bm{P}_{k|k}=\left(\bm{I}-\bm{K}_k{\bm{H}_k}^{\mathrm{T}}\right)\bm{P}_{k|k-1}\triangleq\bm{\Lambda}_k\bm{P}_{k|k-1}
			\label{update}
		\end{aligned}$}
	\end{equation}
	\begin{equation}
		\bm{\xi}_k=\bm{K}_k\left(\bm{y}_k-h\left(\hat{\bm{\chi}}_{k|k-1}\right)\right), 		\hat{\bm{\chi}}_{k|k}=\hat{\bm{\chi}}_{k|k-1}\boxminus\bm{\xi}_k
	\end{equation}
	where $\bm{K}_k$ is the gain matrix at time $t_k$; $\bm{H}_k$ is the measurement matrix; $\bm{N}_k$ is the measurement noise covariance matrix; $h\left(\cdot\right)$ is a function of the state; $\bm{y}_k$ is the measurement of the sensor at $t_k$; $\bm{\Lambda}_k=\left(\bm{I}-\bm{K}_k{\bm{H}_k}^{\mathrm{T}}\right)$ is the correction term \cite{jung2025isolated}; $\bm{\xi}_k$ is the mean of error-state;  $\hat{\bm{\chi}}_{k|k}$ is \textit{a posteriori} estimate with covariance $\bm{P}_{k|k}$; \textit{box minus} $\boxminus$ represent the "minus" on the state manifold \cite{10816004}. During the update steps, the covariances and cross-covariances of all participating agents or sensors will be updated.
}
	\subsection{Efficient Distributed State Estimation via Buffer-Enhanced Covariance Management and Modular Architecture}
	{\label{subb}
	D-GVIO employs a modular framework, where each agent maintains sliding buffers based on sensor frequency. These include: a high-rate IMU buffer $\mathcal{B}^{IMU}=\textrm{Hist}\{\bm{\chi}_k,\bm{P}_k,\bm{F}_k,\bm{G}_k, t_k\}$ that stores core states (states and covariance), state transition matrices, and noise transition matrices; as well as low-rate sensor buffers $\mathcal{B}^{GNSS}=\textrm{Hist}\{\bm{\chi}_k,\bm{P}_k,\bm{\Lambda}_k, t_k\}^{GNSS}$ and $\mathcal{B}^{Visual}=\textrm{Hist}\{\bm{\chi}_k,\bm{P}_k,\bm{\Lambda}_k, t_k\}^{Visual}$ for storing sensor states and correction terms.  $\mathrm{Hist}\{\cdot\}$ denotes a time-horizon sliding buffer that chronologically maintains elements within a fixed duration \cite{jung2025isolated}, with older elements being discarded when exceeding the horizon.
	Buffers are fixed-size and determined by sensor rates: $\mathcal{B}^{IMU}$ is larger due to the higher IMU frequency, while $\mathcal{B}^{GNSS}$ and $\mathcal{B}^{Visual}$ are smaller to reduce the memory burden caused by storing cross-covariances. This strategy maintains constant system maintenance complexity and decouples different sensor states to enable modular processing. 
	
	It is worth noting that D-GVIO adopts a tightly-coupled VIO framework~\cite{8258858}, which increases the dimensionality of the covariance matrix compared to loosely-coupled approaches, leading to higher computational costs. To address this, the covariance segmentation method is adopted to decouple cross-covariances from full-covariance propagation. The full-covariance $\bm{P}_{k|k}$ is partitioned into  core covariance $\bm{P}^C_{k|k}$ and cross-covariance $\bm{P}^{C,S}_{k|k}$. Instead of propagating the full-covariance matrix at each time step, the cross-covariance propagation is deferred and reconstructed in the next update step using stored state transition matrices $\bm{F}_k$ and correction terms $\bm{\Lambda}_k$ in the buffers (see Fig. \ref{pro}). This cross-covariance propagation strategy improves efficiency by avoiding full-covariance propagation and enables modular sensor integration by keeping the core state estimates independent.
	
	Moreover, when agent $A_0$ and others co-observe a landmark or match SLAM features, a collaborative update is triggered on $A_0$. While maintaining consistency typically requires tracking inter-agent cross-covariances, this becomes memory-intensive as the number of agents grows. To address this, D-GVIO adopts the CI algorithm, which enables consistent data fusion without explicitly maintaining inter-agent correlations \cite{11128538}. Specifically, the CI algorithm produces a more conservative covariance estimate than the real estimate one could obtain considering the cross-correlation, ensuring the IEKF remains consistent and avoids overconfidence.
	More importantly, CI algorithm entirely eliminates the need to store and communicate inter-agent cross-covariances, significantly reducing memory and bandwidth requirements. This makes the CI particularly well-suited for resource-constrained multi-agent systems. Furthermore, to achieve low-bandwidth request-response communication, D-GVIO employs VLAD to generate compact binary scene descriptors, which can reduce inter-agent data exchange volume. Each agent sends a RequestUAV message at 10 Hz containing a binary VLAD  descriptor generated from ORB features using a vocabulary of 64 centroids. The RequestUAV message weighs approximately 2.05 $kB$, significantly smaller than a full MessageUAV (approximately 190.7 $kB$).
		\begin{figure}[htbp]
		\centering
		\includegraphics[width=0.35\textwidth]{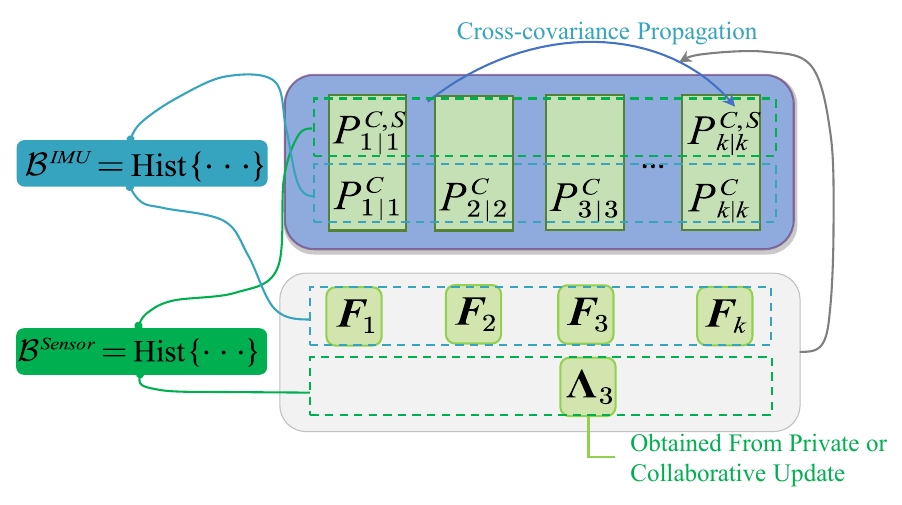}
		\caption{Cross-covariance propagation is delayed until the next update step (at time $t_k$), and is propagated using the state transition matrix and correction term stored in the buffer.}
		\label{pro}
	\end{figure}
}
	\subsection{Efficient Delayed Measurement Handling Strategy Using IEKF}
	{\label{subc}
		In order to enhance real-time performance and better adapt to the distributed nature of multi-agent systems, D-GVIO decouples IMU processing from other sensor measurements (e.g., vision or GNSS) into parallelized independent threads. However, in multithreaded systems, since the processing time of the IMU is often shorter than the update steps of other sensors, computational or communication delays will lead to extensive re-computation steps. In general, when delayed measurements occur (see Fig. \ref{ooo}), the re-propagation strategy \cite{alexander1991state}, which propagates both the covariance and mean from $t_k$ to $t_m$ using the state transition matrices stored in the buffer, can efficiently handle delayed measurements, and this method can handle delayed measurements in private update and cooperative update. The re-propagation steps are as follows:
		\begin{figure}[tbp]
			\centering
			\includegraphics[width=0.45\textwidth]{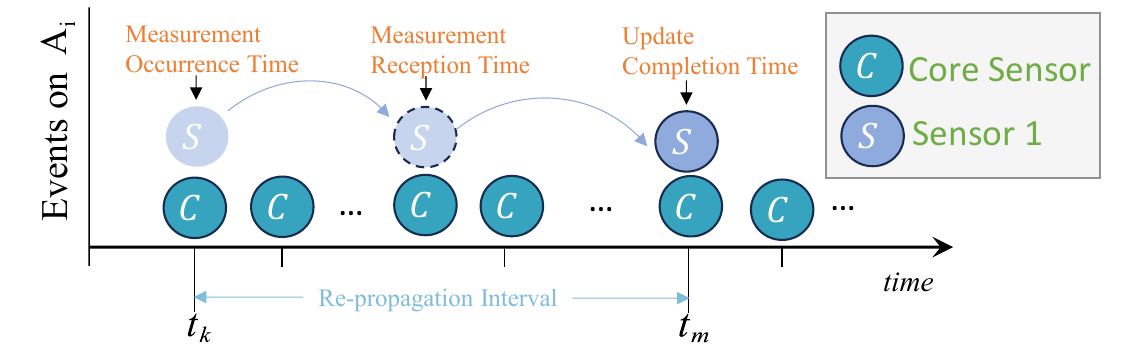}
			\caption{Delayed measurement scenario: Sensor 1 performs update step at time $t_k$. Due to the communication and computational delays, the update step is not completed until time $t_m$.}
			\label{ooo}
		\end{figure}
		\begin{equation}
			\resizebox{0.88\hsize}{!}{$
				\begin{aligned}
					&\bm{P}^{C}_{m|m-1} = 
					\left(\prod_{i=k+1}^m \bm{F}^C_{i}\right)
					\bm{P}^{C}_{k|k}
					\left(\prod_{i=k+1}^m \bm{F}^C_{i}\right)^{\mathrm{T}} + 
					\bm{M}_{m,k+1},\bm{\xi}_m=\left(\prod_{i=k+1}^m \bm{F}^C_{i}\right)\bm{\xi}_k
				\end{aligned}
				$}
			\label{repropagation_Sigma}
		\end{equation}
		\begin{equation}
			\resizebox{0.88\hsize}{!}{$
			\bm{M}_{m,k+1}=\sum^{m-1}_{i=k+1}\bm{F}^C_{i}\bm{G}_{i-1}\bm{Q}_{i-1}\left(\bm{F}^C_{i}\bm{G}_{i-1}\right)^{\mathrm{T}}+\bm{G}_m\bm{Q}_m\bm{G}_m^\mathrm{T}
			$}
		\end{equation}
		where $\bm{M}_{m,k+1}$ is the accumulated process noise from time $t_{k+1}$ to $t_m$.
		Therefore, using the state transition matrices stored in the buffer, the covariance and mean at time $t_m$ can be efficiently computed through (\ref{repropagation_Sigma}).
		
		However, in traditional EKF, $\bm{F}_k = \frac{\partial f}{\partial \bm{\xi}}|{(\hat{\bm{\chi}}_{k|k-1}, u_k)}$ is linearized at the estimated state $\hat{\bm{\chi}}_{k|k-1}$. An initial estimate that deviates from the true state may yield an inaccurate $\bm{F}_k$ , resulting in inconsistency issues. This issue is further exacerbated when delayed measurements occur, as re-propagation of the covariance and mean via (\ref{repropagation_Sigma}) is required. The time-varying $\bm{F}_k$ during re-propagation introduces inconsistency in the covariance propagation path. Although recomputing each $\bm{F}_k$ would be accurate, it is computationally expensive. To address this, the IEKF is adopted for information fusion. In IEKF, the system dynamics are required to be group affine.
		This property guarantees that the left-invariant error or the right-invariant error is autonomous, resulting in their state transition matrix only depend on the control input $u_k$, not on the estimated state $\hat{\bm{\chi}}_{k|k-1}$ \cite{7523335}. According to \cite{fornasier2025equivariant}, the state transition matrices and process noise matrix of L-IEKF and right-invariant EKF (R-LIEKF) are as follows:
\begin{equation}
	\resizebox{0.88\hsize}{!}{$
	\bm{F}^L_k=e^{\bm{A}^L\Delta t},\quad\bm{A}^L=\left[
	\begin{array}{ccc:cc}  
		-\bm{\omega}^b_{ib}\times & \mathbf{0}_{3 \times 3} &\mathbf{0}_{3 \times 3} & \mathbf{I}_{3 \times 3}&\mathbf{0}_{3 \times 3}\\
		-\bm{f}^b_{ib}\times & -\bm{\omega}^b_{ib}\times & \mathbf{0}_{3 \times 3}&\mathbf{0}_{3 \times 3}&\mathbf{I}_{3 \times 3} \\
		\mathbf{0}_{3 \times}&\mathbf{I}_{3 \times}&-\bm{\omega}^b_{ib}\times&\mathbf{0}_{3 \times}&\mathbf{0}_{3 \times}\\
		\hdashline
		&\mathbf{0}_{6 \times 9} & & \multicolumn{2}{c}{\mathbf{0}_{6 \times 6}} 
	\end{array}
	\right]$}
	\label{LIEKF}
\end{equation}
\begin{equation}
	\resizebox{0.88\hsize}{!}{$
	\bm{F}^R_k=e^{\bm{A}^R\Delta t},\quad\bm{A}^R=\left[
	\begin{array}{ccc:cc}  
		\mathbf{0}_{3 \times 3} & \mathbf{0}_{3 \times 3} &\mathbf{0}_{3 \times 3} & \hat{\bm{R}}^w_I&\mathbf{0}_{3 \times 3}\\
		\bm{g}\times & \mathbf{0}_{3 \times 3} & \mathbf{0}_{3 \times 3}&\hat{\bm{v}}^w_I\times\hat{\bm{R}}^w_I&\hat{\bm{R}}^w_I \\
		\mathbf{0}_{3 \times}&\mathbf{I}_{3 \times}&\mathbf{0}_{3 \times 3}&\hat{\bm{p}}^w_I\times\hat{\bm{R}}^w_I&\mathbf{0}_{3 \times}\\
		\hdashline
		&\mathbf{0}_{6 \times 9} & & \multicolumn{2}{c}{\mathbf{0}_{6 \times 6}} 
	\end{array}
	\right]$}
	\label{RIEKF}
\end{equation}
\begin{equation}
	\resizebox{0.88\hsize}{!}{$
	\bm{G}^L_k=\left[
\begin{array}{cc:c}  
	\mathbf{I}_{6 \times 6} & \mathbf{0}_{6 \times 3} & \mathbf{0}_{6 \times 6} \\
	\mathbf{0}_{3 \times 6} & \mathbf{0}_{3 \times 3} & \mathbf{0}_{3 \times 6} \\
	\hdashline
	\multicolumn{2}{c}{\mathbf{0}_{6 \times 9}} & \mathbf{I}_{6 \times 6} 
\end{array}
	\right],\bm{G}^R_k=\left[
	\begin{array}{ccc:c}  
		\hat{\bm{R}}^w_I & \mathbf{0}_{3 \times 3} &\mathbf{0}_{3 \times 3} &\mathbf{0}_{3 \times 6}\\
		\hat{\bm{v}}^w_I\times\hat{\bm{R}}^w_I &\hat{\bm{R}}^w_I & \mathbf{0}_{3 \times 3}&\mathbf{0}_{3 \times 6} \\
		\hat{\bm{p}}^w_I\times\hat{\bm{R}}^w_I&\mathbf{0}_{3 \times 3}&\mathbf{0}_{3 \times 3}&\mathbf{0}_{3 \times 6}\\
		\hdashline
		&\mathbf{0}_{6 \times 9} & & \multicolumn{1}{c}{\mathbf{I}_{6 \times 6}} 
	\end{array}
	\right]
	$}
	\label{G}
\end{equation}
where $\bm{\omega}^b_{ib}$ and $\bm{f}^b_{ib}$ represent the outputs of the gyroscope and accelerometer, respectively; $\bm{g}$ is the gravity vector; $\bm{A}^L$ and $\bm{A}^R$ are the linearized error state matrices of L-IEKF and R-IEKF, respectively; $\bm{G}^L_k$ and $\bm{G}^R_k$ are the process noise matrices of L-IEKF and R-IEKF, respectively; $\Delta t$ is the sampling period.

As shown in (\ref{LIEKF}), (\ref{RIEKF}) and (\ref{G}), $\bm{F}^L_k$ and $\bm{G}^L_k$ are independent of the state estimate, whereas $\bm{F}^R_k$ and $\bm{G}^R_k$ exhibit state dependence. We attribute this to the inclusion of biases $\bm{b}_g$ and $\bm{b}_a$ in the system state, which is defined in the body frame. L-IEKF defines all error states in the body frame, naturally aligning with the bias coordinate system and avoiding state dependence introduced by coordinate transformation. Therefore, state-independent L-IEKF is used in D-GVIO for information fusion.
	}
	\subsection{GNSS Outlier Detection and Culling Strategy}
	{\label{subd}
		GNSS measurements in challenging scenarios (e.g., tunnels, dense urban areas) are prone to outliers, which can severely degrade fusion-based odometry accuracy. However, conventional outlier detection methods like the chi-square test \cite{1104658} suffer from covariance sensitivity and high computational cost (due to covariance  inversion operations). Therefore, a buffer-based position constraint outlier culling (PCOC) algorithm is proposed to overcome these challenges. PCOC estimates velocity $\bm{v}^{GNSS}_k$ from consecutive GNSS positions $\bm{p}^{GNSS}_{k}$ and $\bm{p}^{GNSS}_{k-1}$ and compares it to a dynamic threshold derived from the mean and standard deviation within the GNSS buffer $\mathcal{B}^{GNSS}$. If $\bm{v}^{GNSS}_k$ significantly exceeds the threshold, the GNSS position $\bm{p}^{GNSS}_{k}$ is regarded as an outlier and subsequently removed.
		
		To enhance outlier detection robustness, we propose an entropy-adaptive velocity threshold strategy that jointly exploits GNSS and VIO kinematic information. In this strategy, the local VIO module provides stable velocity estimates $\bm{v}^{VIO}_{k}$, from which we derive the acceleration $|\bm{a}^{VIO}_k| = |\bm{v}^{VIO}_k - \bm{v}^{VIO}_{k-1}|/\Delta t$. The acceleration $|\bm{a}^{VIO}_k|$ quantifies the system's dynamic uncertainty through the lens of information entropy, where higher acceleration implies greater entropy production. This entropy-acceleration relationship follows from the Boltzmann entropy principle $S = k_B \ln \Omega$, with $\Omega$ representing the accessible state volume and $k_B$ is the Boltzmann constant. Specifically, the threshold adjustment factor $k$ is modeled as a function of the normalized velocity $r$ and acceleration $|\bm{a}^{VIO}_k|$ according to the $3\sigma$ criterion:
		\begin{equation}
			\resizebox{0.88\hsize}{!}{$
			V_{threshold}=\mu+k\sigma,\quad k=3+\left(1+r\right)\cdot \mathrm{ln}\left(1+|\bm{a}^{VIO}_k|\right)
			$}
		\end{equation}
		where $V_{threshold}$ is the velocity threshold, $r=min(\bm{v}_k^{VIO}/\bm{v}_k^{emp},1)$ is the normalized velocity with $\bm{v}_k^{emp}$ as the empirical maximum velocity.  The logarithmic term $\ln(1+|\bm{a}^{VIO}_k|)$ encodes the entropy production rate, and the factor $(1+r)$ incorporates velocity-dependent effects. This design ensures $3\sigma$ bound for static scenarios while enabling adaptive threshold relaxation during acceleration. The specific process is depicted in Algorithm \hyperref[Al1]{I}.
		\begin{flushleft}
			\label{Al1}
			\begin{table}[!htbp]
				\resizebox{0.90\linewidth}{!}{
					\begin{tabular}{ll}
						\toprule
						\textbf{Algorithm 1:}  Position Constraint Outlier Culling\\
						\toprule
						\textbf{Input:} GNSS positions $\bm{p}^{GNSS}_{k}$, VIO positions $\bm{p}^{VIO}_{k}$, $\mathcal{B}^{GNSS}$ \\
						GNSS buffer size $W$, Parameters $\beta$ and $\bm{v}^{emp}$\\
						\midrule
						\textbf{1. Preprocessing:} \\
						\quad Compute GNSS velocity squared: \\ \quad\quad ${\bm{v}^{GNSS}_k} = \frac{1}{\Delta t}\sum_{k=1}^3 (\bm{p}^{GNSS}_{k} - \bm{p}^{GNSS}_{k-1})$ \\
						\quad Compute VIO velocity squared: \\ \quad\quad ${\bm{v}^{VIO}_k} = \frac{1}{\Delta t}\sum_{k=1}^3 (\bm{p}^{VIO}_{k} - \bm{p}^{GNSS}_{k-1})$ \\
						\quad Compute the mean and standard deviation: \\
						\quad\quad $\mu = \text{mean}([{\bm{v}^{\text{GNSS}}_{k-W}},{\bm{v}^{\text{GNSS}}_{k-W+1}},...,{\bm{v}^{\text{GNSS}}_{k}}^])$\\
						\quad\quad$\sigma = \text{std}([{\bm{v}^{\text{GNSS}}_{k-W}},{\bm{v}^{\text{GNSS}}_{k-W+1}},...,{\bm{v}^{\text{GNSS}}_{k}}])$ \\
						\textbf{2. Compute velocity threshold:}\\
						\quad $V_{\text{threshold}} = \mu + k \sigma$\\
						\textbf{3. Outlier detection:}\\
						\quad \textbf{if} (${\bm{v}_k^{GNSS}}>V_{threshold}$)  \textbf{then}\\
						\quad \quad $\bm{p}_k^{GNSS}=\textrm{Null}$$\leftarrow$ If it is an outlier, remove it.\\
						\quad \textbf{else}\\
						\quad \quad $\bm{p}_k^{GNSS}=\bm{p}_k^{GNSS}$$\leftarrow$ If it is not an outlier, store it in the $\mathcal{B}^{GNSS}$.\\
						\quad \quad $\mathcal{B}^{GNNS}.push\left(\bm{p}_k^{GNSS}\right)$\\
						\quad \textbf{end if}\\
						\textbf{3. return }$\bm{p}^{GNSS}_k$,$\mathcal{B}^{GNSS}$\\
						\bottomrule
				\end{tabular}}
			\end{table}
		\end{flushleft}
}
\vspace{-20pt}
\section{Experiments}
\label{exp}
	In this section, we evaluate the proposed D-GVIO using both public and self-collected datasets, including eight representative sequences. First, we compare the proposed D-GVIO with state-of-the-art (SOTA) approaches on the public simulation Castle Around dataset \cite{9844233}. Next, we assess the D-GVIO using EKF, L-IEKF, and R-IEKF, along with a comparative analysis between D-GVIO's distributed state estimation approach and centralized EKF on the public S3E Square \cite{10740801} as well as our real-world datasets. In the experiments, the initial yaw is derived from a dual-antenna GNSS system, which calculates the heading by measuring the phase difference of carrier signals between two physically separated antennas. In D-GVIO, the buffer sizes are configured as follows: the IMU buffer $\mathcal{B}^{IMU}$ stores up to 300 states, while the GNSS buffer $\mathcal{B}^{GNSS}$ and visual buffer $\mathcal{B}^{Visual}$ each store up to 10 states. Given the current lack of open-source decentralized GVIO systems, we evaluate the VIO performance of D-GVIO against the state-of-the-art decentralized cooperative systems, COVINS \cite{schmuck2021covins} and X-VIO \cite{9844233}. For GVIO performance, we compare D-GVIO with two representative single-agent systems: IC-GVINS \cite{9961886}, which is based on graph optimization, and InGVIO~\cite{10040716}, which is a filter-based approach. For all three datasets, RTK positioning solutions were adopted as the reference ground truth. All experiments were conducted on a uniform hardware platform (Ubuntu 20.04 LTS, Intel Xeon Silver 4214R, 64GB RAM). 
			\begin{figure*}[htbp]
		\centering
		\includegraphics[width=0.95\textwidth]{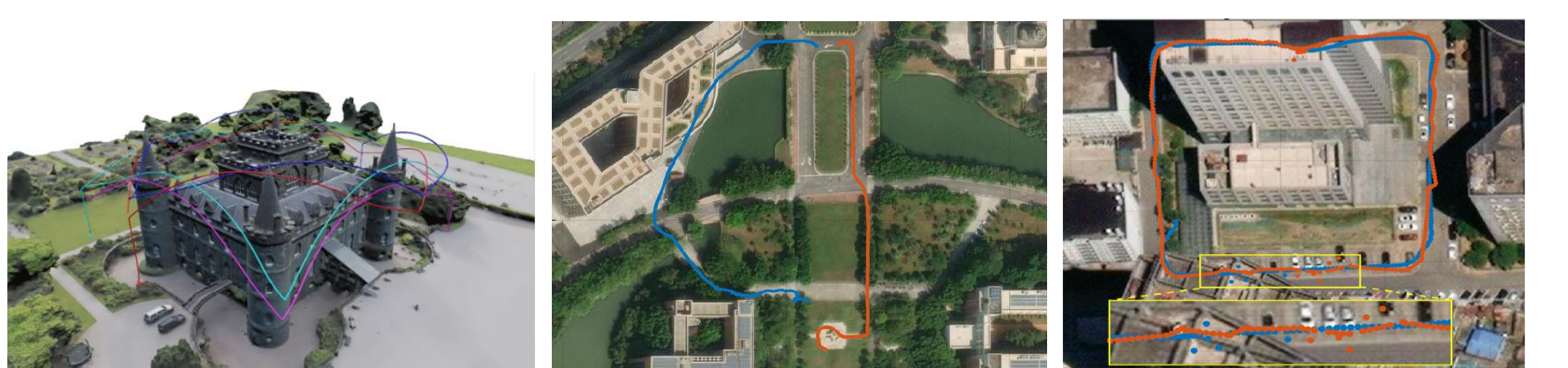}
		\caption{Dataset scenarios (a) "Castle Around" four drones fly around the Inveraray Castle model (https://skfb.ly/6z7Rr) by Andrea Spognetta licensed under Creative Commons Attribution-NonCommercial. (b) "S3E Square" two wheeled robots driving around the square. (c) "our real-world dataset" two wheeled robots driving in parallel around a building. The subplot within the yellow box highlights GNSS outliers.}
		\label{dataset}
	\end{figure*}
	\subsection{Simulation Experiments}
	The Castle Around dataset (see Fig. \ref{dataset}(a)) simulates four UAVs flying cooperatively around a 3D model of Inveraray Castle, following a square trajectory of 220 m. It is generated using the VI-Sensor Simulator (752$\times$480, 30 Hz), with simulated IMU (ADIS16448, 200 Hz) and RTK positioning solution (10 Hz), featuring diverse altitudes and orientations over 30 seconds of flight.
		\subsubsection{Performance Comparison with SOTA Approaches}
	A comprehensive comparison with SOTA approaches is summarized in Tables \ref{tab:ate} and \ref{performance}. As shown in Tables \ref{tab:ate} and \ref{performance}, D-GVIO achieves absolute trajectory error (ATE) comparable to X-VIO, but with significantly lower central processing unit (CPU) and memory usage. Unlike the centralized EKF in X-VIO, D-GVIO propagates only core covariances and employs an efficient delayed-measurement processing strategy, substantially reducing memory and computation costs. The ATE of D-GVIO is slightly higher than that of COVINS. This outcome is expected, as the COVINS system incorporates loop closure detection and global pose and map optimization capabilities. Nevertheless, D-GVIO demonstrates reduced memory usage, lower CPU usage, and lower number of messages exchanged between agents compared to COVINS. This is because the factor graph approach requires storing the poses of historical keyframes, and landmarks. The factor graph optimization process involves iteratively solving nonlinear least-squares problems, resulting in significant memory and computational burdens. 
	Compared with IC-GVINS, D-GVIO maintains similar sub-decimeter ATE but reduces CPU usage by 20-30\% and memory consumption by over 70\%, demonstrating significantly higher efficiency. In contrast, when compared with the filter-based InGVIO, D-GVIO achieves superior ATE accuracy while also exhibiting lower CPU and memory usage.
		\subsubsection{Consistency}
\begin{figure}[h] 
	\centering
	\includegraphics[width=0.42\textwidth]{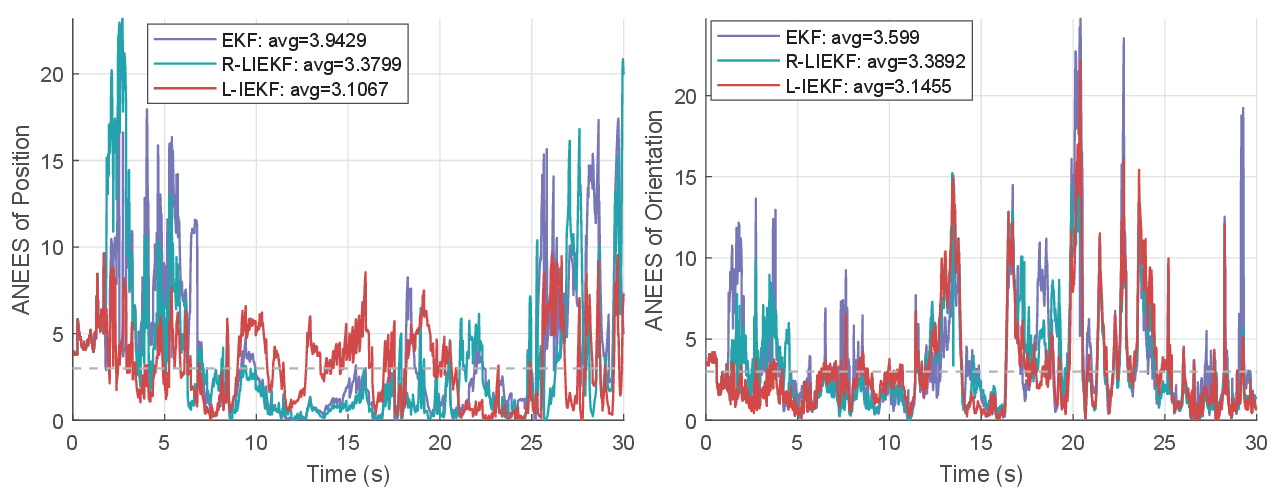}
	\caption{The ANEES of estimated position and orientation using different filters.}
	\label{nees}
\end{figure}

	Normalized Estimation Error Squared (NEES) is an important metric for evaluating the consistency of filter estimation performance, used to determine whether the filter is overly optimistic or conservative. Fig. \ref{nees} demonstrates the average normalized estimated error squared (ANEES) of the pose estimation by D-GVIO using L-IEKF, R-IEKF, and traditional EKF over 10 simulation runs. Due to space constraints, we have selected the ANEES results of UAV 1 for demonstration. Note that the ANEES of the position and orientation should be on average 3. In the experimental results, the traditional EKF yields more optimistic uncertainty estimates compared to other two filters. Meanwhile, the L-IEKF demonstrates the position and orientation ANEES closer to the reference value than the R-IEKF. These experimental results confirms the importance of keeping the state transition matrix independent of system estimates in D-GVIO.
			\begin{table}[h]
		\centering
		\caption{ATE (meter) comparison with the SOTA on the Castle Around dataset.}
		\label{tab:ate}
		\setlength{\tabcolsep}{4pt}  
		\resizebox{0.90\linewidth}{!}{  
			\begin{tabular}{llSSSSSS}
				\toprule
				\textbf{Datasets} & \textbf{Agent} & 
				\textbf{Ours} & \textbf{X-VIO} & \textbf{COVINS} & \textbf{Ours} & \textbf{InGVIO}& \textbf{IC-GVINS} \\
				& & \textbf{(VIO)} & \textbf{(VIO)} & \textbf{(VIO)} & \textbf{(GVIO)} & \textbf{(GVIO)} & \textbf{(GVIO)}\\
				\midrule
				\multirow{4}{*}{Castle Around} 
				& UAV 1 & 0.67 & 0.70 & 0.51 & \rm{0.07} &0.08& \textbf{0.06} \\
				& UAV 2 & 0.74 & 0.73 & 0.63 &0.11& \rm{0.07} & \textbf{0.06} \\
				& UAV 3 & 0.82 & 0.82 & 0.76 & \textbf{0.08} &0.13& \rm{0.11} \\
				& UAV 4 & 0.79 & 0.83 & 0.80 & \textbf{0.07} &0.09& \rm{0.10} \\
				\bottomrule
		\end{tabular}}
	\end{table}
	\begin{table}[h]
		\centering
		\caption{Algorithm performance comparison with the SOTA on the Castle Around dataset.}
		\label{performance}
		\setlength{\tabcolsep}{3pt}  
		\resizebox{0.9\linewidth}{!}{  
			\begin{tabular}{llS[table-format=4.0]S[table-format=3.2]S[table-format=1.3]}
				\toprule
				\textbf{Algorithm} & \textbf{Agent} & 
				\textbf{N. of Messages sent} & \textbf{Max CPU (\%)} & \textbf{Max Mem (MiB)} \\
				\midrule
				\multirow{4}{*}{Ours} 
				& UAV 1   & \textbf{1042} & \textbf{135.66} & \textbf{81.21} \\
				& UAV 2   & \textbf{1068} & \textbf{142.01} & \textbf{89.31} \\
				& UAV 3   & \textbf{1231} & 167.75 & \textbf{95.22} \\
				& UAV 4   & \rm{1256} & \textbf{171.68} & \textbf{92.31} \\
				\midrule
				\multirow{5}{*}{COVINS} 
				& UAV 1   & 3568 & 200.10 & \rm{657.89} \\
				& UAV 2   & 3576 & 205.21 & \rm{648.48} \\
				& UAV 3   & 3833 & 199.31 & \rm{701.31} \\
				& UAV 4   & 3842 & 210.21 & \rm{722.65} \\
				\midrule
				\multirow{5}{*}{X-VIO} 
				& UAV 1   & $\rm{1046}$ & 208.47 & \rm{95.56} \\
				& UAV 2   & $\rm{1055}$ & 211.28 & \rm{102.85} \\
				& UAV 3   & $\rm{1267}$ & 211.28 & \rm{112.42} \\
				& UAV 4   & \textbf{1253} & 211.28 & \rm{115.29} \\
				\midrule
\multirow{5}{*}{InGVIO} 
& UAV 1   & \multicolumn{1}{c}{--} & 145.35 &\rm{92.35} \\
& UAV 2   & \multicolumn{1}{c}{--} & 157.22 & \rm{94.15} \\
& UAV 3   & \multicolumn{1}{c}{--} & \textbf{161.21} & \rm{97.29} \\
& UAV 4   & \multicolumn{1}{c}{--} & 175.68 & \rm{93.87} \\
				\midrule
				\multirow{5}{*}{IC-GVINS} 
				& UAV 1   & \multicolumn{1}{c}{--} & 180.47 & \rm{305.56} \\
				& UAV 2   & \multicolumn{1}{c}{--} & 195.32 & \rm{312.40} \\
				& UAV 3   & \multicolumn{1}{c}{--} & 185.48 & \rm{311.14} \\
				& UAV 4   & \multicolumn{1}{c}{--} & 188.01 & \rm{315.59} \\
				\bottomrule
		\end{tabular}}
	\end{table}
	\subsection{Real World Experiments}
	In the simulation experiments, we conduct comprehensive comparisons and analyses of D-GVIO and SOTA approaches. To further validate the performance of D-GVIO and the improvements of its submodules, we conduct experiments on two sets of real-world datasets: the open-source S3E square dataset and a self-collected dataset.
	The open-source S3E square dataset (Fig. \ref{dataset}(b)) features two wheeled robots (Robot 1 and Robot 2) following square trajectories, with synchronized 10Hz image data and RTK positioning solution.
	Our real-world dataset were collected in a high-rise building-dense area of Wuhan University. This dataset was collected by two remotely controlled wheeled robots (Robot 3 and Robot 4) equipped with industrial-grade cameras, GNSS receivers/antenna, and IMU sensor. Due to severe multipath effects caused by the high-rise buildings, GNSS positioning accuracy was significantly degraded in this environment. This dataset can be used to validate the robustness of the D-GVIO system and the performance of its outlier detection module. The GNSS localization results for both robots are shown in Fig. \ref{dataset}(c). 
				 	\begin{figure}[!htbp]
		\centering
		\includegraphics[width=0.45\textwidth]{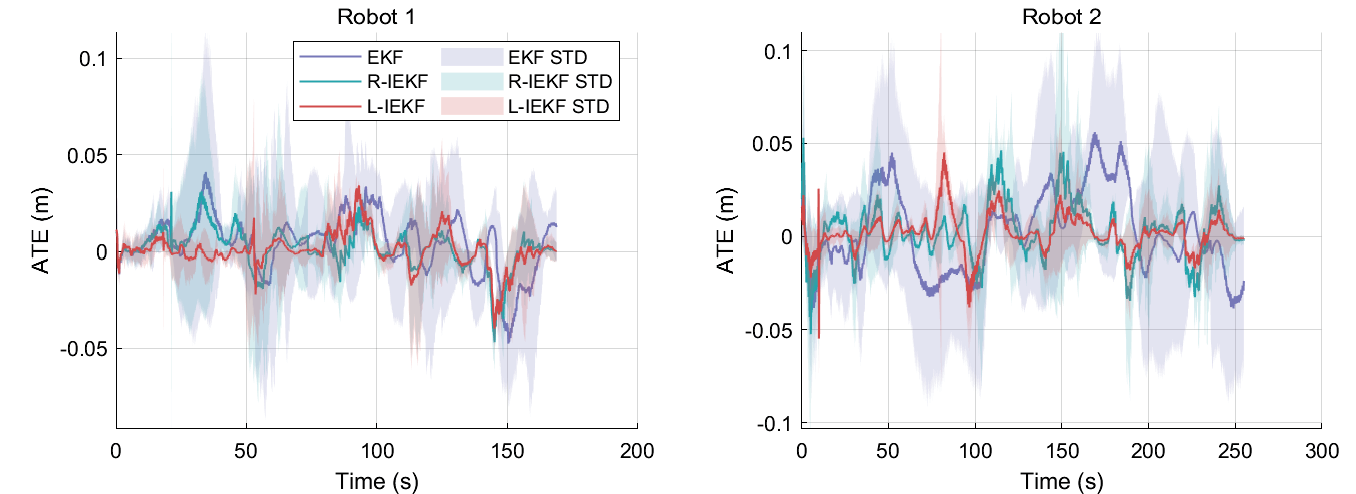}
		\caption{The ATE of D-GVIO using different filters on the S3E Square dataset.}
		\label{ate}
	\end{figure}
		 		\begin{figure}[!htbp]
		\centering
		\includegraphics[width=0.45\textwidth]{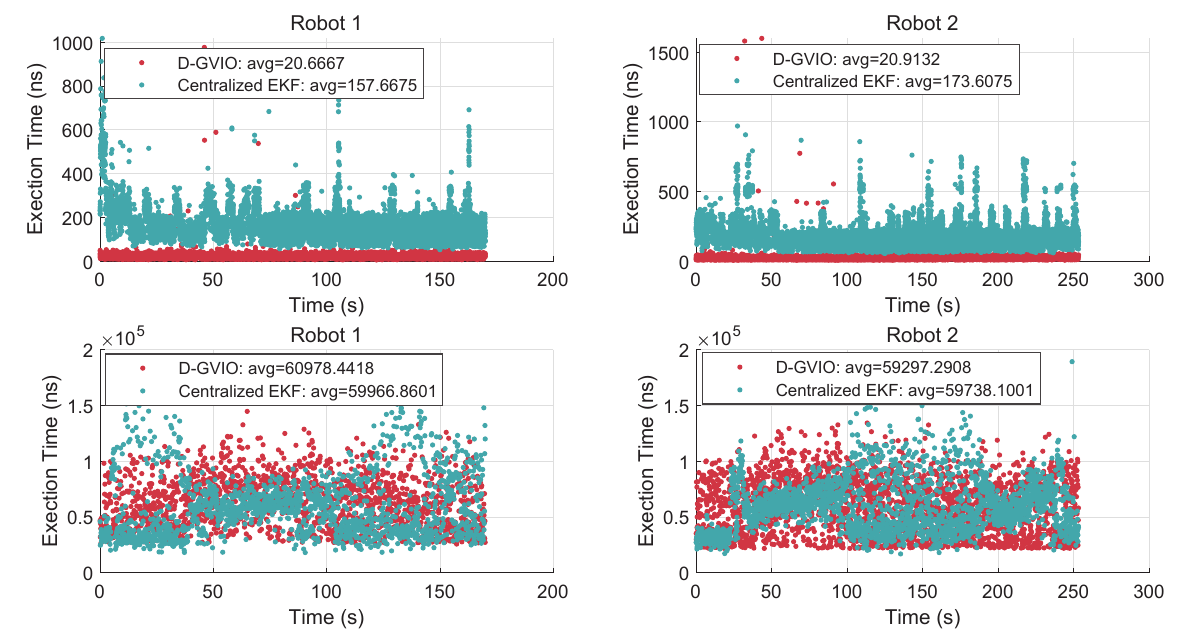}
		\caption{The execution time of D-GVIO and centralized EKF. The legends inform about the average execution time for each robot. The two plots in the first row and second rows show the execution time of propagation and vision-related private and collaborative measurement update steps for two robots, respectively.}
		\label{time}
	\end{figure}
	\subsubsection{Positioning Accuracy and Computational Efficiency}
	 Fig. \ref{ate} demonstrates the ATE of D-GVIO with EKF, L-IEKF, and R-IEKF. In the experimental results for both robots, the traditional EKF exhibits larger ATE compared to other two filters. Meanwhile, the L-IEKF demonstrates a smaller ATE and significantly more stable error curves. These experimental results are consistent with the results reflected by the ANEES for the traditional EKF, L-IEKF, and R-IEKF in our simulation experiments (see Fig. \ref{nees}). Additionally, Fig. \ref{time} shows the execution time of propagation and private/collaborative update steps in D-GVIO compared to the centralized EKF. In propagation, D-GVIO is significantly faster due to the covariance segmentation strategy, which only maintains the core covariance during propagation. In updates, D-GVIO incurs a slight time increase from cross-covariance propagation. However, given the significant reduction in propagation time and the more modular handling of sensors in D-GVIO, this modest increase in update time is acceptable.
	 \begin{figure}[!htbp]
	 	\centering
	 	\includegraphics[width=0.45\textwidth]{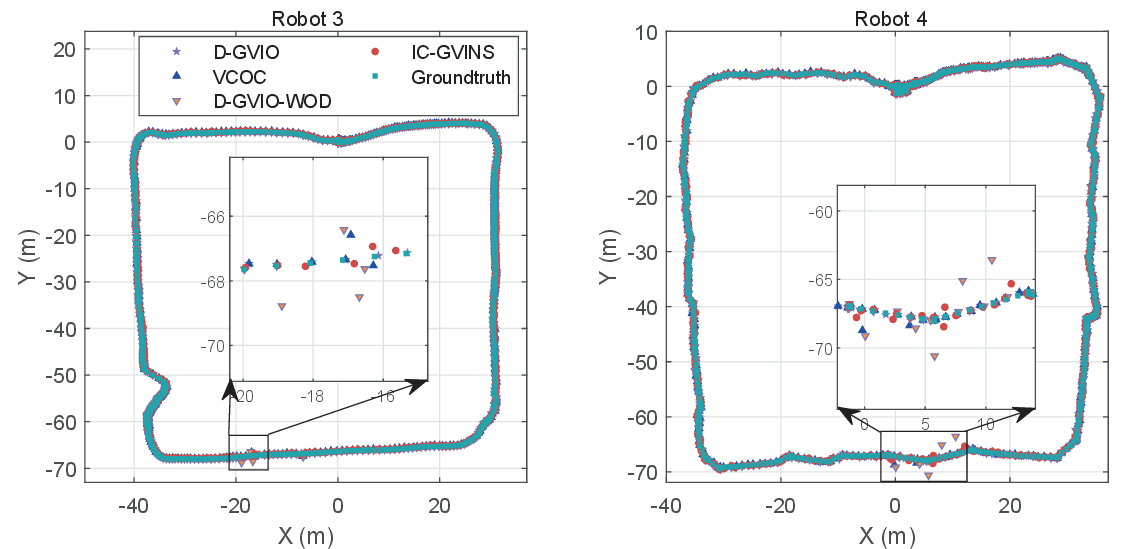}
	 	\caption{Trajectories estimate results for Robot 3 and Robot 4 in the real-world dataset.}
	 	\label{trjcom}
	 \end{figure}
	\subsubsection{GNSS Outlier Detection}
		As shown in Fig. \ref{trjcom}, in GNSS-challenged environments, D-GVIO demonstrates better robustness to GNSS anomalies compared to D-GVIO without the outlier detection module (D-GVIO-WOD), effectively preventing outliers from degrading positioning accuracy and achieving superior localization performance compared to IC-GVINS. Moreover, compared with the computationally efficient velocity constraint outlier culling (VCOC) algorithm \cite{10477234}, D-GVIO demonstrates higher sensitivity to GNSS outliers. As shown in the two subfigures of Fig. \ref{trjcom}, D-GVIO achieves more accurate elimination of GNSS outliers, whereas VCOC is less sensitive to smaller outliers. This is because unlike VCOC, which relies on empirical values for outlier rejection, our method dynamically adjusts the threshold based on the vehicle's acceleration, thereby accounting for the dynamic nature of the driving process. These results demonstrate that our method effectively detects and removes GNSS outliers, thereby avoiding their impact on positioning accuracy.
\FloatBarrier
\section{Conclusion}
{This paper proposes D-GVIO, a decentralized GVIO system featuring a buffer-driven architecture. Unlike conventional approaches, D-GVIO decouples core states and other sensor states via a modular buffer design, significantly reducing computational and memory burdens. Furthermore, D-GVIO adopts a buffer-based re-propagation strategy with IEKF to efficiently handle delayed measurements, as well as an adaptive GNSS outlier detection mechanism to improve robustness in  GNSS-challenged environments. Evaluations on three independent datasets show that D-GVIO achieves higher computational efficiency, lower memory usage, and better accuracy compared to SOTA approaches. Its efficiency and reliability make it well-suited for resource-constrained platforms in multi-agent collaborative localization. Future work will focus on dynamic collaborative networks to enhance robustness and accuracy in large-scale systems.}

		\bibliographystyle{IEEEtran}
\bibliography{referenceooo}
\vspace{-30 mm}

\end{document}